# On Globular T-Spherical Fuzzy (G-TSF) Sets with Application to G-TSF Multi-Criteria Group Decision-Making

**Miin-Shen Yang[1,*], Yasir Akhtar[1], Mehboob Ali[2]**

[1]Department of Applied Mathematics, Chung Yuan Christian University, Chung-Li, Taoyuan 32023, Taiwan

[2] Government Degree College for Boys Sakwar Gilgit, Gilgit-Baltistan, Pakistan

[*]E-mail: msyang@cycu.edu.tw

**Abstract**

In this paper, we give the concept of Globular T-Spherical Fuzzy (G-TSF) Sets (G-TSFSs) as an innovative extension of T-Spherical Fuzzy Sets (TSFSs) and Circular Spherical Fuzzy Sets (C-SFSs). G-TSFSs represent membership, indeterminacy, and non-membership degrees using a globular/sphere bound that can offer a more accurate portrayal of vague, ambiguous, and imprecise information. By employing a structured representation of data points on a sphere with a specific center and radius, this model enhances decision-making processes by enabling a more comprehensive evaluation of objects within a flexible region. Following the newly defined G-TSFSs, we establish some basic set operations and introduce fundamental algebraic operations for G-TSF Values (G-TSFVs). These operations expand the evaluative capabilities of decision-makers, facilitating more sensitive decision-making processes in a broader region. To quantify a similarity measure (SM) between GTSFVs, the SM is defined based on the radius of G-TSFSs. Additionally, Hamming distance and Euclidean distance are introduced for G-TSFSs. We also present theorems and examples to elucidate computational mechanisms. Furthermore, we give the G-TSF Weighted Average (G-TSFWA) and G-TSF Weighted Geometric (G-TSFWG) operators. Leveraging our proposed SM, a Multi-Criteria Group Decision-Making (MCGDM) scheme for G-TSFSs, named G-TSF MCGDM (G-TSFMCGDM), is developed to address group decision-making problems. The applicability and effectiveness of the proposed G-TSFMCGDM method are demonstrated by applying it to solve the selection problem of the best venue for professional development training sessions in a firm. The analysis results affirm the suitability and utility of the proposed method for resolving MCGDM problems, establishing its effectiveness in practical decision-making scenarios.

## 1. Introduction

Zadeh [32] first introduced the concept of fuzzy sets (FSs) as a way to deal with uncertain situations. FSs assign membership degrees to elements of a set in the interval [0, 1], and they had various applications in different fields (see [21, 14, 24]). Afterwards, there are many extensions of FSs in the literature. Atanassov's [9] work on intuitionistic FSs (IFSs) was a significant advancement, as it expanded FSs by introducing the degrees of membership (DoM) $\phi(x)$ and non-membership (DoN) $\psi(x)$ with the condition $0 \le \phi(x) + \psi(x) \le 1$. IFSs has been applied in various fields, such as (see [18, 23, 6]). While Atanassov's construction of IFSs is highly regarded, decision makers often face constraints in allocating values due to the condition on $\phi(x)$ and $\psi(x)$. In some cases, the sum of their DoM exceeds 1, leading to limitations in IFSs. To address this, Yager [27] established the concept of Pythagorean FSs (PyFSs), where DoM and DoN are assigned with the condition $0 \le \phi^2(x) + \psi^2(x) \le 1$, and then Yager [28] extended PyFSs to q-rung orthopair FSs (qROFSs) with the condition $0 \le \phi^q(x) + \psi^q(x) \le 1$ for $q$ being a positive integer. These Yager's extensions provide a more flexible approach to representing DoM, and DoN overcoming the limitations of FSs and IFSs. These PyFSs and q-ROFSs have been applied in many areas (see [33, 11, 29, 2]). On the other hand, Cuong [16] proposed picture fuzzy sets (PFSs) to address the limitations of Atanassov's IFSs, specially the situations where thoughts are not restricted to "yes" or "no" but also include abstention or refusal. PFSs define elements in a triplet representing membership degree (DoM), indeterminacy degree (DoI) and non-membership degree (DoN) $(\phi(x), \chi(x), \psi(x))$ subject to the condition $0 \le \phi(x) + \chi(x) + \psi(x) \le 1$. PFSs extend IFSs into the three ways of memberships with DoM, DoI and DoN which offer a more comprehensive representation of human decision-making processes. Occasionally, the total of the DoM, DoI and DoN exceeds 1, leading to limitations in the application of PFSs. To address this, Mahmood et. al. [22] proposed the framework of spherical FSs (SFSs) by establishing the domain of the degrees $\phi(x)$, $\chi(x)$, and $\psi(x)$ to the condition $0 \le \phi^2(x) + \chi^2(x) + \psi^2(x) \le 1$. In other words, the field of PFSs is limited and the framework of SFSs strengthens the idea of PFSs by increasing the space of DoM, DoI and DoN. However, SFSs only generalize PFSs to some extents. To overcome this limitation, a new structure of T-spherical FSs (TSFSs) was introduced by Ullah et .al. [25], which enables the assignment of any value to $\phi$, $\chi$ and $\psi$ in the interval [0,1] with a condition $0 \leq \phi^n(x) + \chi^n(x) + \psi^n(x) \leq 1$, provided that n is an integer that is equal to or larger than 1. For example, if $\phi = 0.8$, $\chi = 0.5$, and $\psi = 0.4$, then the sum of squares of their membership degrees exceeds 1, but by taking cube of $\phi$, $\chi$ and $\psi$, their sum becomes less than or equal to one. This demonstrates that TSFSs are a more effective technique than SFSs. Consequently, TSFSs have the capability to address situations in which the frameworks of FSs, IFSs, PyFSs, qROFSs, PFSs, and SFSs fall short. There were various applications of TSFSs. Wu et. al. [26] a novel divergence measure within the TSFSs structure is suggested by leveraging the benefits of Jensen-Shannon divergence, termed as TSFSJS distance. Garg et. al. [17] gave the power aggregation operators for TSFSs with their properties and special cases of these operators. Abid et. al. [1] suggested some similarity measures based on TSFSs with applications to pattern recognition.

Another notable extension of IFSs is proposed by Atanassov [10], introducing a circular representation for each point in IFSs. This prolongation is known as circular IFSs (C-IFSs). In this concept, a circle is employed to illustrate the ambiguity associated with DOM and DoN within the set. Specifically, a circle centered at a duo of non-negative real numbers whose sum is less than one, encapsulates the DoM and DoN of all elements within C-IFSs. C-IFSs offer a more nuanced approach to adjusting DoM and DoN, effectively indicating degrees of uncertainty. Atanassov and Marinov [8] then introduced distance measures tailored for C-IFSs. Building upon this, Boltürk and Kahraman [12] introduced the idea of interval-valued representations, while Alkan and Kahraman [5] proposed the circular intuitionistic fuzzy (C-IF) TOPSIS method as a pandemic hospital location selection. Further research has explored various aspects of C-IFSs. Khan et. al. [20] introduced divergence measures and their practical applications within C-IFSs. Yusoff et. al. [31] gave C-IF ELECTRE III model for group decision analysis, and Chen [15] proposed distance metrics for C-IFSs. Furthermore, Bozyigit et. al. [13] extended C-IFSs to the concept of circular PyFSs (C-PyFSs), which represents the idea of DoM and DoN in the terms of circles, subject to the condition $0 \leq \phi^2(x) + \psi^2(x) \leq 1$, where they also considered T-norms and T-conorms with weighted geometric and arithmetic aggregation for C-PyFSs. Khan et. al. [19] then expanded C-PyFSs to disc PyFSs, and Ali and Yang [3]gave C-PyF Hamacher aggregation operators (AOs) with application in the assessment of goldmines. Subsequently, Yusoff et. al. [30] extended C-PyFSs to circular qROFSs (C-qROFSs), and Ali and Yang [4] gave Dombi AOs on C-qROFSs.

Recently, Ashraf et. al. [7] introduced the notion of the so-called circular spherical fuzzy sets (C-SFSs), which represents the idea of DoM, DoI and DoN in term of a sphere with the constraint $0 \le \phi^2(x) + \chi^2(x) + \psi^2(x) \le 1$. They also proposed the Sugeno-Weber operation on C-SFSs to combine multiple sources of information or multiple fuzzy sets into a single output. However, the concept of C-SFSs has its limitations, particularly in scenarios involving the summation of squares of the DOM, DOI and DoN exceeds 1. *i.e.* $0 \le \phi^2(x) + \chi^2(x) + \psi^2(x) > 1$. For instance, when the values with $\phi$ = 0.9, $\chi$ = 0.7, and $\psi$ = 0.6 are considered, the summation of their membership degrees' squares is more than 1. In such instances, the framework of C-SFSs, along with its structural norms and rules, becomes inapplicable. This identified research gap gives us the motivation to propose a new framework of Globular T-spherical fuzzy sets (G-TSFSs) in this paper.

Thus, we introduce the innovative concept of G-TSFSs which should be more comprehensive and robust in overcoming the limitations of C-SFSs and specific cases, including C-PyFSs and C-IFSs. To illustrate by elevating the power of DoM, DoI, and DoN to the "*t*" in the constraints of C-SFSs, we gain the flexibility to assign values within the [0, 1] range that ensure their sum falls within the unit interval [0, 1], where *t* is a positive integer determined by the discretion of decision-makers. For example, if we have $\phi$ = 0.9, $\chi$ = 0.7, and $\psi$ = 0.6 with $0.9^2 + 0.7^2 + 0.6^2 > 1$, then they are not in C-SFSs. However, if we take *t* as 5, then we have $0 \le 0.9^5 + 0.7^5 + 0.6^5 = 0.836 \le 1$, i.e. the sum of their grades remains within [0, 1]. Hence, G-TSFSs can serve as the focal point of our research, and the major contributions of our proposed framework are described as follows:

- Introduction of G-TSFSs as an innovative extension of most existing FS models.
- Representation of DOM, DOI and DoN using a sphere, offering a more precise representation of vague and ambiguous information.

- Establishment of basic set operations and fundamental algebraic operations for G-TSFSs for G-TSF values.
- Proposed similarity measures based on the radius of G-TSFSs as well as including Hamming distance and Euclidean distance measures.
- Theoretical elucidation and numerical examples provided to illustrate computational mechanisms involved in G-TSFSs.
- Introduction of G-TSF weighted average aggregation (G-TSFWA) and G-TSF weighted geometric aggregation (G-TSFWG) operators for G-TSFSs.
- Formation of a MCGDM framework specifically for G-TSFSs, named G-TSF multi-criteria group decision-making (G-TSFMCGDM).
- Application of G-TSFMCGDM to solve the selection problem of the best venue for professional development training sessions in a firm, demonstrating its applicability and effectiveness in practical decision-making scenarios.

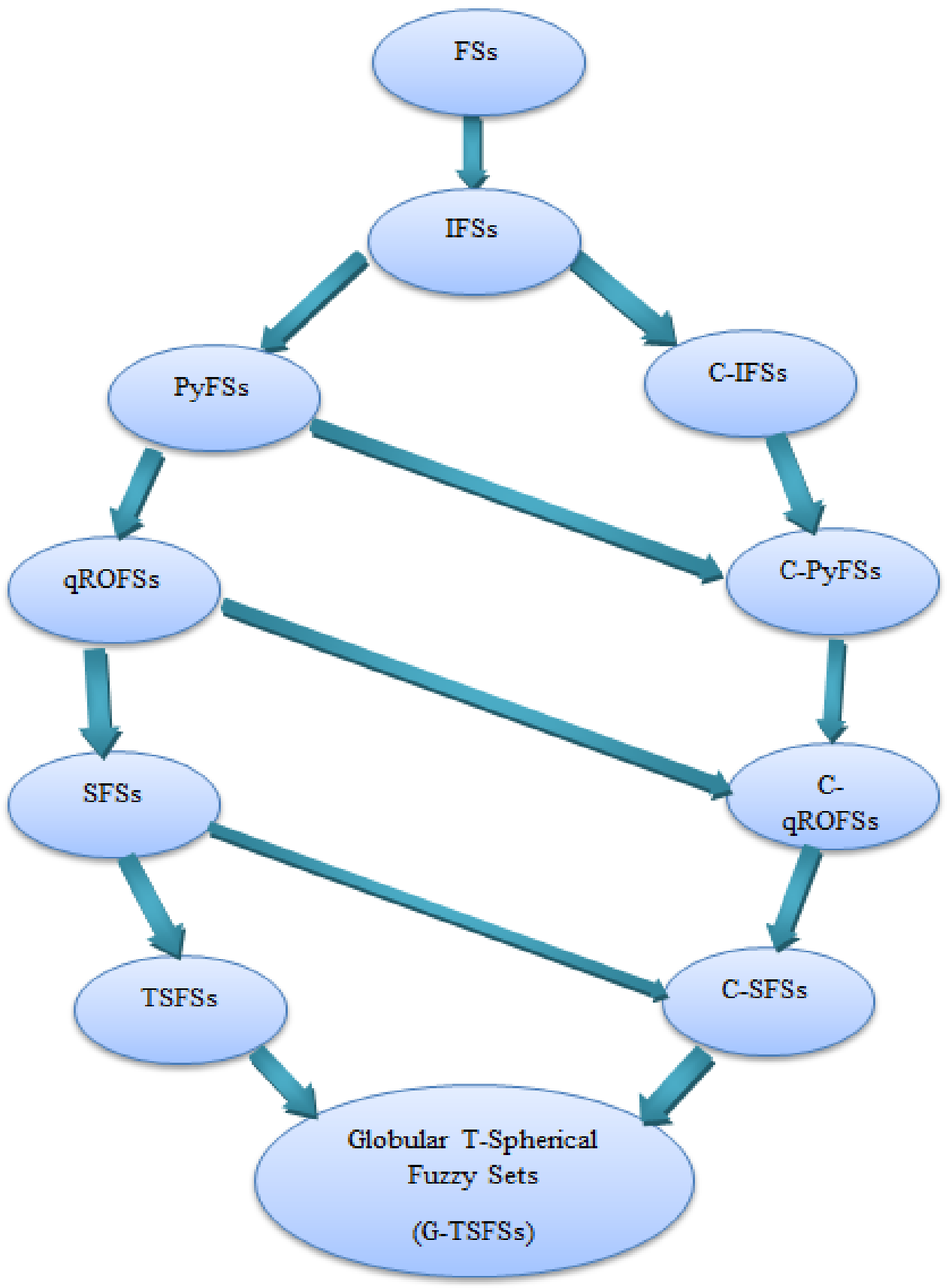

**Fig. 1**. Diagrammatical representation of our proposed model.

The remainder of this paper is structured as follows: In Section 2, we revisit fundamental definitions integral to our study, encompassing various extensions of the fuzzy framework such as IFSs, PyFSs, C-IFSs, C-PyFSs, SFSs, C-SFSs, and TSFSs. In Section 3, we present the groundbreaking concept of G-TSFSs, outlining their fundamental set operations and algebraic functions. This section further explores the introduction of Hamming distance and Euclidean distance measures, accompanied by a detailed discussion on the score function and accuracy function specifically crafted for our proposed sets. Moving on to Section 4, we provide a demonstration of our newly proposed weighted aggregation operators tailored specifically for G-TSFSs. In Section 5, we introduce the formulation of the G-TSF MCGDM (G-TSFMCGDM) model based on the cosine similarity measure. Within this framework, we define a cosine similarity measure specifically tailored for G-TSF values to quantify the extent of similarity among them. Employing this novel similarity measure, we establish a comprehensive MCGDM methodology within the context of a G-TSF environment. To enhance understanding, a practical example is presented. Section 6 outlines the conclusions of the study.

## 2. Preliminaries

In this section, we revisit the fundamental ideas of various types of FSs. We also review the circular framework of IFSs and extensions. We assume that X denotes a non-empty set as a universe of discourses.

**Definition 1**. [9]. An IFS $A$ in X is characterized by the form with $A=\{\langle x,\phi_A(x),\psi_A(x)\rangle : x\in X\}$, where the functions $\phi_A : X\to[0,1]$ and $\psi_A : X\to[0,1]$ present DoM and DoN, respectively, for $x\in X$, under a requirement $0\le\phi_A(x)+\psi_A(x)\le 1$.

**Definition 2** . [10]. A C-IFS $A_r$ in $X$ is define as $A_r=\{\langle x,\phi_A(x),\psi_A(x);r\rangle | x\in X\}$, where $\phi_A : X\to[0,1]$ and $\psi_A : X\to[0,1]$ represent DoM and DoN, respectively, with the condition $0\le\phi_A(x)+\psi_A(x)\le 1$ and $r\in[0,1]$ ensures that the sum of the DoM and DoN of each element does not exceed 1. The radius $r$ signifies the imprecise area around the coordinate formed by DoM and DoN. Each element in C-IFSs is demonstrated by a circle with the center $(\phi_A(x),\psi_A(x))$ and a radius $r$.

**Definition 3**. [27]. A $P_yFS$ $\breve{Y}$ in $X$ is defined as $\breve{Y}=\left\{\langle x,\phi_{\breve{Y}}(x),\psi_{\breve{Y}}(x)\rangle : x\in X\right\}$, where $\phi_{\breve{Y}}:X\to[0,1]$ and $\psi_{\breve{Y}}:X\to[0,1]$ represents DoM and DoN, respectively, with the condition $0\le\phi_{\breve{Y}}^2+\psi_{\breve{Y}}^2\le 1$. Further, the pair $\alpha=\langle\phi_{\breve{Y}},\psi_{\breve{Y}}\rangle$ is called a $P_yF$ Value ($P_yFV$).

**Definition 4**. [13]. A C-$P_yFS$ on a universal set $X$ with $r\in[0,1]$ is define as $\breve{C}_r=\left\{\langle x,\phi_{\breve{C}}(x),\psi_{\breve{C}}(x);r\rangle | x\in X\right\}$, where $\phi_{\breve{C}}:X\to[0,1]$ and $\psi_{\breve{C}}:X\to[0,1]$ represent DoM and DoN, respectively, and they satisfy the condition $0\le\phi_{\breve{C}}^2(x)+\psi_{\breve{C}}^2(x)\le 1$. The point $(\phi_{\breve{C}}(x),\psi_{\breve{C}}(x))$ on the plane represents the center of the circle with a radius value of $r$.

**Definition 5**. [16]. A Picture Fuzzy Set (PFS) $\breve{P}$ in a universal set $X$ is defined as the form $\breve{P}=\left\{\langle x,\phi_{\breve{P}}(x),\chi_{\breve{P}}(x),\psi_{\breve{P}}(x)\rangle : x\in X\right\}$, where $\phi_{\breve{P}}:X\to[0,1]$, $\chi_{\breve{P}}:X\to[0,1]$, and $\psi_{\breve{P}}:X\to[0,1]$ signify DoM, DoI, and DoN for $x\in X$, respectively, and $\phi_{\breve{P}},\chi_{\breve{P}}$ and $\psi_{\breve{P}}$ satisfy the condition $0\le\phi_{\breve{P}}(x)+\chi_{\breve{P}}(x)+\psi_{\breve{P}}(x)\le 1$. The triplet $\alpha=\langle\phi_{\breve{P}},\chi_{\breve{P}},\psi_{\breve{P}}\rangle$ is called a picture fuzzy value (PFV).

**Definition 6**. [22]. A SFS $\breve{S}$ in a universal set $X$ is defined as $\breve{S}=\left\{\langle x,\phi_{\breve{S}}(x),\chi_{\breve{S}}(x),\psi_{\breve{S}}(x)\rangle : x\in X\right\}$, where $\phi_{\breve{S}}:X\to[0,1]$, $\chi_{\breve{S}}:X\to[0,1]$, and $\psi_{\breve{S}}:X\to[0,1]$ indicate DoM, DoI, and DoN of each $x\in X$, respectively. Furthermore, these $\phi_{\breve{S}},\chi_{\breve{S}}$ and $\psi_{\breve{S}}$ satisfy $0\le\phi_{\breve{S}}^2(x)+\chi_{\breve{S}}^2(x)+\psi_{\breve{S}}^2(x)\le 1$ for all $x\in X$. Then the triplet $\alpha=\langle\phi_{\breve{S}},\chi_{\breve{S}},\psi_{\breve{S}}\rangle$ is called a SF value (SFV).

**Definition 7.** [25]. A T-SFS $\breve{T}$ in a universe of discourse X takes the form $\breve{T}=\left\{\langle x,\phi_{\breve{T}}(x),\chi_{\breve{T}}(x),\psi_{\breve{T}}(x)\rangle : x\in X\right\}$ where $\phi_{\breve{T}}:X\to[0,1]$, $\chi_{\breve{T}}:X\to[0,1]$, and $\psi_{\breve{T}}:X\to[0,1]$ represent DoM, DoI, and DoN of each $x\in X$, respectively. Additionally, it holds that $0\le\phi_{\breve{T}}^q+\chi_{\breve{T}}^q+\psi_{\breve{T}}^q\le 1$ for all $q\in Z$. Then, the triplet $\alpha=\langle\phi_{\breve{T}},\chi_{\breve{T}},\psi_{\breve{T}}\rangle$ is called a T-SFV.

**Definition 8**. [7]. A circular SFS (C-SFS) $\breve{P}$ in a universal set $X$ is defined as the form $\breve{S}_r=\left\{\langle x,\phi_{\breve{S}}(x),\chi_{\breve{S}}(x),\psi_{\breve{S}}(x);r\rangle : x\in X\right\}$, where $\phi_{\breve{S}}:X\to[0,1]$, $\chi_{\breve{S}}:X\to[0,1]$, and $\psi_{\tilde{S}} : X \to [0,1]$ signify DoM, DoI, and DoN for $x \in X$ , respectively, and $\phi_{\tilde{S}}, \chi_{\tilde{S}}$ and $\psi_{\tilde{S}}$ satisfy the condition $0 \le \phi_{\tilde{S}}^2(x) + \chi_{\tilde{S}}^2(x) + \psi_{\tilde{S}}^2(x) \le 1$ for all $x \in X$ .

### 3. The Proposed Globular T-Spherical Fuzzy Sets

In this section, we present an innovative and advanced concept, called Globular T-Spherical Fuzzy Sets (G-TSFSs), which serves as a novel extension encompassing IFSs, CIFSs, PyFSs, CPyFSs, PFSs, CPFSs, SFSs, CSFSs, and TSFSs. It allows decision makers to articulate and handle uncertainty, vagueness, impressions, and indeterminacy in a more nuanced manner. G-TSFSs achieve this by introducing the use of a sphere around a central point of DoM, DoI, and DoN on a broader conceptual space. These three components provide a more refined and comprehensive way of capturing and quantifying different aspects of uncertainty. The broader space and environment are provided by a sphere centered in each point of DoM, DoI, and DoN on the three-dimensional space. G-TSFSs can offer decision makers a more holistic perspective, and so decision makers can conduct a more thorough evaluation during the decision-making process. The assessment mechanism benefits from this enhanced representation of uncertainty, leading to improved accuracy and overall performance. The resulting from a more effective decision-making framework can produce superior outcomes.

**Definition 9.** Let $X$ be a universal set of discourses. A Globular T-Spherical Fuzzy Set (G-TSFS) $G_r$ in $X$ is defined as

$$G_r = \{\langle x, \phi_G(x), \chi_G(x), \psi_G(x); r \rangle : x \in X\} \tag{1}$$

where $\phi_G : X \to [0,1]$, $\chi_G : X \to [0,1]$, and $\psi_G : X \to [0,1]$ represent DoM, DoI, and DoN, respectively, under the conditions $0 \le \phi_G^t(x) + \chi_G^t(x) + \psi_G^t(x) \le 1$ for $t$ being any positive integer, and $r \in [0,1]$ is the radius of the sphere centered in the point $\langle \phi_G(x), \chi_G(x), \psi_G(x) \rangle$ on the space. The central point $\langle \phi_G(x), \chi_G(x), \psi_G(x) \rangle$ is derived by using the values of DoM, DoI, and DoN in TSFVs under consideration.

Each element in a TSFS is characterized as a particular location under the triangular fuzzy interpretation. However, in G-TSFSs, all of elements in it are visualized as a sphere characterized by its center denoted by $\langle \phi_G(x), \chi_G(x), \psi_G(x) \rangle$, and a radius denoted by $r$. A TSFS can be expressed in the form of G-TSFSs as $G_0 = \{\langle x, \phi_G(x), \chi_G(x), \psi_G(x); 0 \rangle : x \in X\}$. Thus, the concept of G-TSFSs is the generalization of the notion of TSFSs. However, a G-TSFS with $r > 0$ cannot be handled by using these norms in TSFSs. Thus, norms in G-TSFSs need to be re-constructed, and it is supposed that these norms in G-TSFSs should be broader and superior as compared to those in TSFSs.

**Definition 10.** Let $\alpha = \langle \phi_\alpha, \chi_\alpha, \psi_\alpha \rangle$ with $\phi_\alpha, \chi_\alpha, \psi_\alpha \in [0,1]$ be a TSFV such that $\phi_\alpha^t + \chi_\alpha^t + \psi_\alpha^t \le 1$ and let $r_\alpha \in [0,1]$ be the radius in G-TSFVs. Then, the quartet $\alpha_r = \langle \phi_\alpha, \chi_\alpha, \psi_\alpha; r_\alpha \rangle$ is called a G-TSF value (G-TSFV).

**Remark 1.** In Definition 9, we have the following special cases:

- When setting $t = 2$, G-TSFSs transform into C-SFSs.
- When setting $r = 0$, G-TSFSs transform into TSFSs.
- When setting $\chi_\alpha = 0$, G-TSFSs transform into C-qROFSs
- When setting $t = 2$ and $\chi_\alpha = 0$, G-TSFSs transform into C-PyFSs.
- When setting $t = 1$ and $\chi_\alpha = 0$, G-TSFSs transform into C-IFSs.

In the followings, we present the computational mechanism for determining the central point $\langle \phi(x), \chi(x), \psi(x) \rangle$ and the radius $r$ of the sphere using the membership grades of DoM, DoI, and DoN. We also demonstrate the process of conversion from TSFVs to G-TSFVs with the help of $\langle \phi(x), \chi(x), \psi(x) \rangle$ and $r$ to represent G-TSFSs in terms of the sphere. Let $\{\langle \phi_1, \chi_1, \psi_1 \rangle, \langle \phi_2, \chi_2, \psi_2 \rangle ..., \langle \phi_n, \chi_n, \psi_n \rangle\}$ be a family of TSFVs. We consider the point $\langle \bar{\phi}, \bar{\chi}, \bar{\psi} \rangle$ defined as:

$$\langle \bar{\phi}, \bar{\chi}, \bar{\psi} \rangle = \left\langle \sqrt[t]{\frac{\sum_{j=1}^{n} \phi_j^t}{n}}, \sqrt[t]{\frac{\sum_{j=1}^{n} \chi_j^t}{n}}, \sqrt[t]{\frac{\sum_{j=1}^{n} \psi_j^t}{n}} \right\rangle \tag{2}$$

Then, we can show that $\langle \bar{\phi}, \bar{\chi}, \bar{\psi}; r \rangle$ for $r \in [0,1]$ becomes to be a G-TSFV. Essentially, the radius $r$ of the sphere corresponds to the maximum distance from the central point $\langle \bar{\phi}, \bar{\chi}, \bar{\psi} \rangle$ to any point on its surface within the collection of TSFVs constituting a G-TSFV. This can be expressed mathematically as follows:

$$r = \min\left\{ \max_{1 \le j \le n} \sqrt{(\bar{\phi}^t - \phi_j^t)^2 + (\bar{\chi}^t - \chi_j^t)^2 + (\bar{\psi}^t - \psi_j^t)^2}, 1 \right\} \tag{3}$$

**Theorem 1.** Let $\{\langle \phi_1, \chi_1, \psi_1 \rangle, \langle \phi_2, \chi_2, \psi_2 \rangle, \ldots, \langle \phi_n, \chi_n, \psi_n \rangle\}$ be a family of TSFVs. Then $\langle \bar{\phi}, \bar{\chi}, \bar{\psi}; r \rangle$ is a G-TSFV where $\bar{\phi} = \sqrt[t]{\sum_{j=1}^{n} \phi_j^t \Big/ n}$, $\bar{\chi} = \sqrt[t]{\sum_{j=1}^{n} \chi_j^t \Big/ n}$, $\bar{\psi} = \sqrt[t]{\sum_{j=1}^{n} \psi_j^t \Big/ n}$ and $r$ is defined as in Eq. (3).

Proof: We have that

$$\bar{\phi}^t + \bar{\chi}^t + \bar{\psi}^t = \left( \sqrt[t]{\sum_{j=1}^{n} \phi_j^t \Big/ n} \right)^t + \left( \sqrt[t]{\sum_{j=1}^{n} \chi_j^t \Big/ n} \right)^t + \left( \sqrt[t]{\sum_{j=1}^{n} \psi_j^t \Big/ n} \right)^t = \frac{\sum_{j=1}^{n} \phi_j^t}{n} + \frac{\sum_{j=1}^{n} \chi_j^t}{n} + \frac{\sum_{j=1}^{n} \psi_j^t}{n}$$

$$= \frac{\sum_{j=1}^{n} \phi_j^t + \sum_{j=1}^{n} \chi_j^t + \sum_{j=1}^{n} \psi_j^t}{n} = \frac{\sum_{j=1}^{n} (\phi_j^t + \chi_j^t + \psi_j^t)}{n} \le \frac{\sum_{j=1}^{n} (1)}{n} = 1.$$

That is, $\bar{\phi}^t + \bar{\chi}^t + \bar{\psi}^t \le 1$.

It is clear that $0 \le r = \min\left\{ \max_{1 \le j \le n} \sqrt{(\bar{\phi}^t - \phi_j^t)^2 + (\bar{\chi}^t - \chi_j^t)^2 + (\bar{\psi}^t - \psi_j^t)^2}, 1 \right\} \le 1$. Thus $\langle \bar{\phi}, \bar{\chi}, \bar{\psi}; r \rangle$ is a G-TSFV. ■

**Example 1:** Let us present the following collection of TSFVs: $\{\langle 0.4, 0.6, 0.65 \rangle, \langle 0.2, 0.7, 0.4 \rangle, \langle 0.6, 0.5, 0.6 \rangle, \langle 0.5, 0.3, 0.4 \rangle\}$, $\{\langle 0.6, 0.3, 0.1 \rangle, \langle 0.5, 0.5, 0.3 \rangle, \langle 0.8, 0.2, 0.1 \rangle, \langle 0.7, 0.2, 0.3 \rangle\}$ and $\{\langle 0.6, 0.5, 0.3 \rangle, \langle 0.78, 0.4, 0.3 \rangle, \langle 0.7, 0.3, 0.2 \rangle, \langle 0.65, 0.1, 0.4 \rangle\}$. By applying Eqs. (2) and (3), we obtain the relevant G-TSFVs as: $\langle 0.42, 0.52, 0.52; 0.30 \rangle$, $\langle 0.65, 0.3, 0.2; 0.26 \rangle$ and $\langle 0.68, 0.32, 0.3; 0.57 \rangle$. In general, a G-TSFS is basically the family of G-TSFVs. The data in this example regarding TSFVs is portrayed in Fig. 2(left), and the generalization of TSFVs to G-TSFVs is shown in Fig. 2(right).

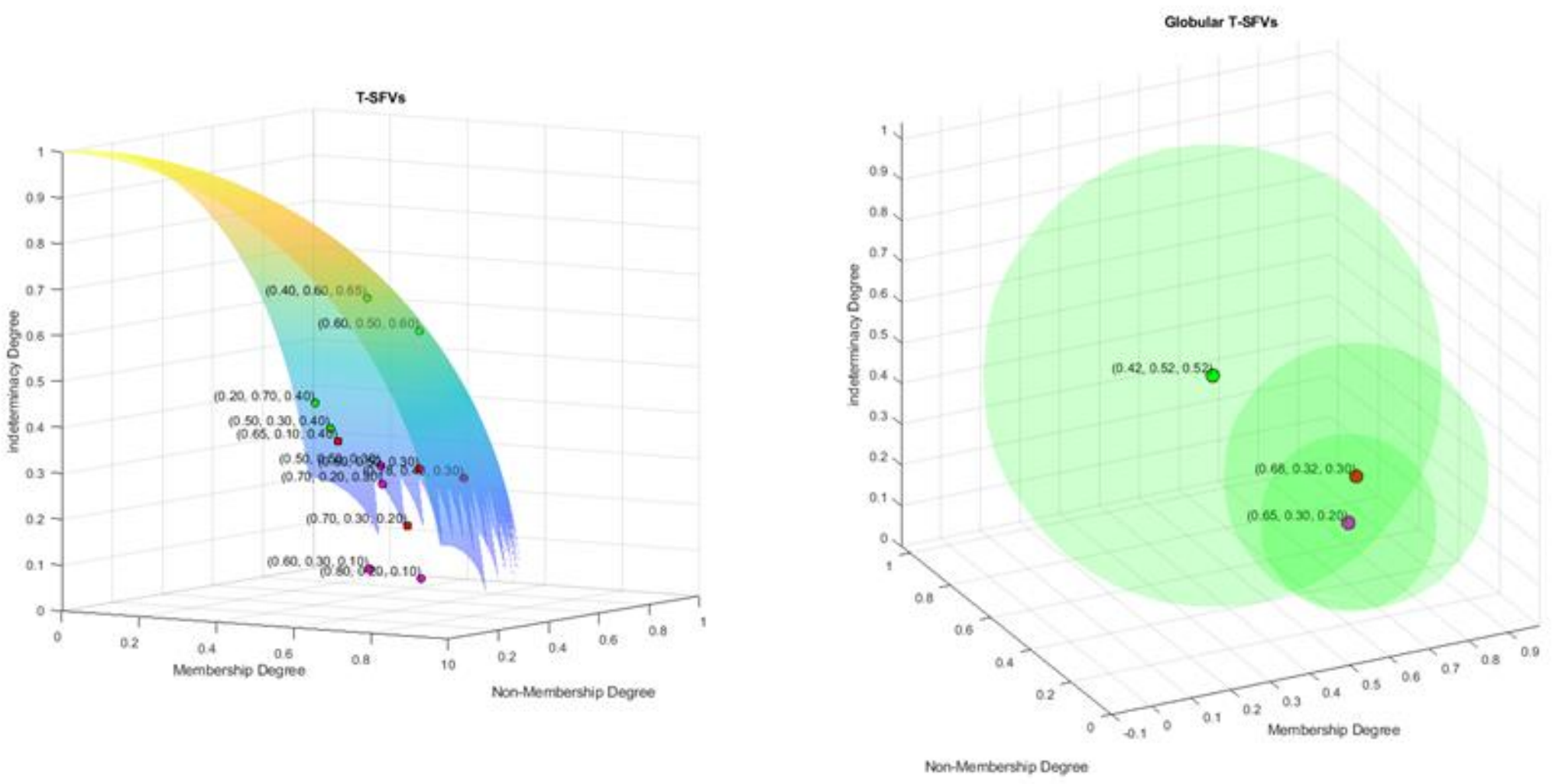

Fig. 2. (left): TSFVs; (right): G-TSFVs

We next give some fundamental operations for our proposed G-TSFSs as follows.

**Definition 11.** Let us have the two G-TSFSs $\widehat{M}_r$ and $\widehat{N}_s$ in $X$ with $\widehat{M}_r = \left\{ \left\langle x, \phi_{\widehat{M}}(x), \chi_{\widehat{M}}(x), \psi_{\widehat{M}}(x); r \right\rangle : x \in X \right\}$ and $\widehat{N}_s = \left\{ \left\langle x, \phi_{\widehat{N}}(x), \chi_{\widehat{N}}(x), \psi_{\widehat{N}}(x); s \right\rangle : x \in X \right\}$. Then we define some basic set operations on G-TSFSs as:

(i) $\widehat{M}_r \subset \widehat{N}_s$ iff $r \le s$ and $\phi_{\widehat{M}}(x) \le \phi_{\widehat{N}}(x), \chi_{\widehat{M}}(x) \le \chi_{\widehat{N}}(x)$ , and $\psi_{\widehat{M}}(x) \ge \psi_{\widehat{N}}(x)$

(ii) $\widehat{M}_r = \widehat{N}_s$ iff $r = s$ and $\phi_{\widehat{M}}(x) = \phi_{\widehat{N}}(x), \chi_{\widehat{M}}(x) = \chi_{\widehat{N}}(x)$ , and $\psi_{\widehat{M}}(x) = \psi_{\widehat{N}}(x)$

(iii) $\widehat{M}_r^c = \left\{ \left\langle x, \psi_{\widehat{M}}(x), \chi_{\widehat{M}}(x), \phi_{\widehat{M}}(x); r \right\rangle : x \in X \right\}$

(iv) $\widehat{M}_r \cup_{\min} \widehat{N}_s = \left\{ x, \max(\phi_{\widehat{M}}(x), \phi_{\widehat{N}}(x)), \min(\chi_{\widehat{M}}(x), \chi_{\widehat{N}}(x)), \min(\psi_{\widehat{M}}(x), \psi_{\widehat{N}}(x)); \min(r,s) : x \in X \right\}$.

(v) $\widehat{M}_r \cup_{\max} \widehat{N}_s = \left\{ x, \max(\phi_{\widehat{M}}(x), \phi_{\widehat{N}}(x)), \min(\chi_{\widehat{M}}(x), \chi_{\widehat{N}}(x)), \min(\psi_{\widehat{M}}(x), \psi_{\widehat{N}}(x)); \max(r,s) : x \in X \right\}$.

(vi) $\widehat{M}_r \cap_{\min} \widehat{N}_s = \left\{ x, \min(\phi_{\widehat{M}}(x), \phi_{\widehat{N}}(x)), \min(\chi_{\widehat{M}}(x), \chi_{\widehat{N}}(x)), \max(\psi_{\widehat{M}}(x), \psi_{\widehat{N}}(x)); \min(r,s) : x \in X \right\}$.

(vii) $\widehat{M}_r \cap_{\max} \widehat{N}_s = \left\{ x, \min(\phi_{\widehat{M}}(x), \phi_{\widehat{N}}(x)), \min(\chi_{\widehat{M}}(x), \chi_{\widehat{N}}(x)), \max(\psi_{\widehat{M}}(x), \psi_{\widehat{N}}(x)); \max(r,s) : x \in X \right\}$.

We give a numerical example to demonstrate the above fundamental operations.

**Example 2:** Let the universal set be X={$x_1$, $x_2$, $x_3$}, and the two G-TSFSs on X are: $\widehat{M}_{0.3} = \{ \langle x_1, 0.4, 0.6, 0.65; 0.3 \rangle, \langle x_2, 0.2, 0.7, 0.4; 0.3 \rangle, \langle x_3, 0.6, 0.5, 0.6; 0.3 \rangle \}$ and $\widehat{N}_{0.7} = \{\langle x_1, 0.8, 0.4, 0.5; 0.7\rangle, \langle x_2, 0.5, 0.5, 0.3; 0.7\rangle, \langle x_3, 0.8, 0.4, 0.5; 0.7\rangle\}$. Then, we have that $\widehat{M}_{0.3} \subset \widehat{N}_{0.7}$. We also obtain $\widehat{M}_{0.3}^{\,c} = \{\langle x_1, 0.65, 0.6, 0.4; 0.3\rangle, \langle x_2, 0.4, 0.7, 0.2; 0.3\rangle, \langle x_3, 0.6, 0.5, 0.6; 0.3\rangle\}$; $\widehat{M}_{0.3} \cup_{\min} \widehat{N}_{0.7} = \{\langle x_1, 0.8, 0.4, 0.5; 0.3\rangle, \langle x_2, 0.5, 0.5, 0.3; 0.3\rangle, \langle x_3, 0.8, 0.4, 0.5; 0.3\rangle\}$; $\widehat{M}_{0.3} \cup_{\max} \widehat{N}_{0.7} = \{\langle x_1, 0.8, 0.4, 0.5; 0.7\rangle, \langle x_2, 0.5, 0.5, 0.3; 0.7\rangle, \langle x_3, 0.8, 0.4, 0.5; 0.7\rangle\}$; $\widehat{M}_{0.3} \cap_{\min} \widehat{N}_{0.7} = \{\langle x_1, 0.4, 0.4, 0.65; 0.3\rangle, \langle x_2, 0.2, 0.5, 0.4; 0.3\rangle, \langle x_3, 0.6, 0.4, 0.6; 0.3\rangle\}$ and

$\widehat{M}_{0.3} \cap_{\max} \widehat{N}_{0.7} = \{\langle x_1, 0.4, 0.4, 0.65; 0.7\rangle, \langle x_2, 0.2, 0.5, 0.4; 0.7\rangle, \langle x_3, 0.6, 0.4, 0.6; 0.7\rangle\}$.

**Theorem 2.** Let $\widehat{M}_r = \{\langle x, \phi_{\widehat{M}}(x), \chi_{\widehat{M}}(x), \psi_{\widehat{M}}(x); r\rangle : x \in X\}$ and $\widehat{N}_s = \{\langle x, \phi_{\widehat{N}}(x), \chi_{\widehat{N}}(x), \psi_{\widehat{N}}(x); s\rangle : x \in X\}$ be the two G-TSFSs in $X$. Then, we have the following results:

(1) $(\widehat{M}_r \cap_{\min} \widehat{N}_s)^c = \widehat{M}_r^{\,c} \cup_{\min} \widehat{N}_s^{\,c}$.

(2) $(\widehat{M}_r \cap_{\max} \widehat{N}_s)^c = \widehat{M}_r^{\,c} \cup_{\max} \widehat{N}_s^{\,c}$

(3) $(\widehat{M}_r \cup_{\min} \widehat{N}_s)^c = \widehat{M}_r^{\,c} \cap_{\min} \widehat{N}_s^{\,c}$

(4) $(\widehat{M}_r \cup_{\max} \widehat{N}_s)^c = \widehat{M}_r^{\,c} \cap_{\max} \widehat{N}_s^{\,c}$

Proof: We first prove (1) as follows.

L.H.S. $= (\widehat{M}_r \cap_{\min} \widehat{N}_s)^c$

$$= \left(\{\langle x, \phi_{\widehat{M}}(x), \chi_{\widehat{M}}(x), \psi_{\widehat{M}}(x); r\rangle : x \in X\} \cap_{\min} \{\langle x, \phi_{\widehat{N}}(x), \chi_{\widehat{N}}(x), \psi_{\widehat{N}}(x); s\rangle : x \in X\}\right)^c$$

$$= \{\langle x, \min(\phi_{\widehat{M}}(x), \phi_{\widehat{N}}(x)), \min(\chi_{\widehat{M}}(x), \chi_{\widehat{N}}(x)), \max(\psi_{\widehat{M}}(x), \psi_{\widehat{N}}(x)); \min(r, s)\rangle : x \in X\}^c$$

$$= \{\langle x, \max(\psi_{\widehat{M}}(x), \psi_{\widehat{N}}(x)), \min(\chi_{\widehat{M}}(x), \chi_{\widehat{N}}(x)), \min(\phi_{\widehat{M}}(x), \phi_{\widehat{N}}(x)); \min(r, s)\rangle : x \in X\}.$$

R.H.S= $\widehat{M}_r^{\,c} \cup_{\min} \widehat{N}_s^{\,c}$

$$= \left(\{\langle x, \phi_{\widehat{M}}(x), \chi_{\widehat{M}}(x), \psi_{\widehat{M}}(x); r\rangle : x \in X\}\right)^c \cup_{\min} \left(\{\langle x, \phi_{\widehat{N}}(x), \chi_{\widehat{N}}(x), \psi_{\widehat{N}}(x); s\rangle : x \in X\}\right)^c$$

$$= \{\langle x, \psi_{\widehat{M}}(x), \chi_{\widehat{M}}(x), \phi_{\widehat{M}}(x); r\rangle : x \in X\} \cup_{\min} \{\langle x, \psi_{\widehat{N}}(x), \chi_{\widehat{N}}(x), \phi_{\widehat{N}}(x); s\rangle : x \in X\}$$

$$=\left\{\left\langle x,\max(\psi_{\widehat{M}}(x),\psi_{\widehat{N}}(x)),\min(\chi_{\widehat{M}}(x),\chi_{\widehat{N}}(x)),\min(\phi_{\widehat{M}}(x),\phi_{\widehat{N}}(x));\min(r,s)\right\rangle : x\in X\right\}.$$

Thus, L.H.S. = R.H.S., and so we prove (1). For (2) ~ (4), the proofs are similar as (1). ■

In the followings, we introduce some basic algebraic operators on G-TSFVs. These can be used in developing various measures, operators, and functions for the proposed framework.

**Definition 12.** For two G-TSFVs $\breve{Y}_r$ and $\breve{Z}_s$ with $\breve{Y}_r=\left\langle \phi_{\breve{Y}},\chi_{\breve{Y}},\psi_{\breve{Y}};r\right\rangle$ and $\breve{Z}_s=\left\langle \phi_{\breve{Z}},\chi_{\breve{Z}},\psi_{\breve{Z}};s\right\rangle$, the following algebraic operators on G-TSFVs are defined:

(i) $\breve{Y}_r \oplus_{\min} \breve{Z}_s=\langle \sqrt[t]{\phi_{\breve{Y}}^t+\phi_{\breve{Z}}^t-\phi_{\breve{Y}}^t.\phi_{\breve{Z}}^t},\chi_{\breve{Y}}.\chi_{\breve{Z}},\psi_{\breve{Y}}.\psi_{\breve{Z}};\min(r,s)\rangle$

(ii) $\breve{Y}_r \oplus_{\max} \widehat{N}_s=\langle \sqrt[t]{\phi_{\breve{Y}}^t+\phi_{\breve{Z}}^t-\phi_{\breve{Y}}^t.\phi_{\breve{Z}}^t},\chi_{\breve{Y}}.\chi_{\breve{Z}},\psi_{\breve{Y}}.\psi_{\breve{Z}};\max(r,s)\rangle$

(iii) $\breve{Y}_r \otimes_{\min} \breve{Z}_s=\langle \phi_{\breve{Y}}.\phi_{\breve{Z}},\chi_{\breve{Y}}.\chi_{\breve{Z}},\sqrt[t]{\psi_{\breve{Y}}^t+\psi_{\breve{Z}}^t-\psi_{\breve{Y}}^t.\psi_{\breve{Z}}^t};\min(r,s)\rangle$

(vi) $\breve{Y}_r \otimes_{\max} \breve{Z}_s=\langle \phi_{\breve{Y}}.\phi_{\breve{Z}},\chi_{\breve{Y}}.\chi_{\breve{Z}},\sqrt[t]{\psi_{\breve{Y}}^t+\psi_{\breve{Z}}^t-\psi_{\breve{Y}}^t.\psi_{\breve{Z}}^t};\max(r,s)\rangle$

### 3.1. Globular TSF Ranking

In this subsection, we give the procedure for evaluating G-TSFVs. This evaluation process allows us for the comparison and prioritization in G-TSFVs by taking into account of their individual properties and characteristics.

**Definition 13**: Let $\alpha_r=\left\langle \phi_\alpha,\chi_\alpha,\psi_\alpha;r_\alpha\right\rangle$ be a G-TSFV. The score function and accuracy function for the G-TSFVs are defined respectively as:

Score function: $$\widehat{\gamma}(\alpha)=\frac{1}{2}\left(\phi_\alpha^t-\chi_\alpha^t-\psi_\alpha^t+r_\alpha\left(2\sigma-1\right)\right) \tag{4}$$

where $\widehat{\gamma}(\alpha)\in[-1,1]$ and $\sigma\in[0,1]$

Accuracy function: $$\widehat{\tau}(\alpha)=\phi_\alpha^t+\chi_\alpha^t+\psi_\alpha^t \tag{5}$$

where, $\widehat{\tau}(\alpha)\in[0,1]$.

**Definition 14**: Let $\breve{Y}_r$ and $\breve{Z}_s$ be two G-TSFVs with $\breve{Y}_r=\left\langle \phi_{\breve{Y}},\chi_{\breve{Y}},\psi_{\breve{Y}};r\right\rangle$ and $\breve{Z}_s=\left\langle \phi_{\breve{Z}},\chi_{\breve{Z}},\psi_{\breve{Z}};s\right\rangle$, respectively. Then, we define a ranking mechanism for G-TSFVs on the basis of the newly introduced score function (4) and accuracy function (5) as follows.

- If $\widehat{\gamma}(\breve{Y})>\widehat{\gamma}(\breve{Z})$, then $\breve{Y}>\breve{Z}$

- If $\hat{\gamma}(\breve{Y}) < \hat{\gamma}(\breve{Z})$, then $\breve{Y} < \breve{Z}$

If $\hat{\gamma}(\breve{Y}) = \hat{\gamma}(\breve{Z})$, then

- If $\hat{\tau}(\breve{Y}) > \hat{\tau}(\breve{Z})$, then $\breve{Y} > \breve{Z}$
- If $\hat{\tau}(\breve{Y}) < \hat{\tau}(\breve{Z})$, then $\breve{Y} < \breve{Z}$
- If $\hat{\tau}(\breve{Y}) = \hat{\tau}(\breve{Z})$, then $\breve{Y} \approx \breve{Z}$.

We mention that, in Definition 9, we had defined a G-TSFS $G_r$ in $X$ as $G_r = \left\{\left\langle x, \phi_G(x), \chi_G(x), \psi_G(x); r\right\rangle : x \in X\right\}$. In fact, we may define a more general type of G-TSFS in $X$ with $\breve{G}_r = \left\{\left\langle x, \phi_{\breve{G}}(x), \chi_{\breve{G}}(x), \psi_{\breve{G}}(x); r(x)\right\rangle : x \in X\right\}$ where the radius $r(x)$ is considered to depend on $x \in X$. Of course, if $r(x) = r$ (a constant) for all $x \in X$, then it becomes the same G-TSFS as defined in Definition 9. We next define some distance measures on these general G-TSFSs. Assume that the two G-TSFSs $\breve{A}_r$ and $\breve{B}_s$ are with $\breve{A}_r = \left\{\left\langle x, \phi_{\breve{A}_r}(x), \chi_{\breve{A}_r}(x), \psi_{\breve{A}_r}(x); r(x)\right\rangle : x \in X\right\}$, and $\breve{B}_s = \left\{\left\langle x, \phi_{\breve{B}_s}(x), \chi_{\breve{B}_s}(x), \psi_{\breve{B}_s}(x); s(x)\right\rangle : x \in X\right\}$.

**Definition 15**. The Hamming distance for two G-TSFSs $\breve{A}_r$ and $\breve{B}_s$ under each $x \in X$ is defined as:

$$H_1(\breve{A}_r, \breve{B}_s) = \frac{1}{2}\left(\left|r(x) - s(x)\right| + \frac{1}{2}\left(\left|\phi^t_{\breve{A}_r}(x) - \phi^t_{\breve{B}_s}(x)\right| + \left|\chi^t_{\breve{A}_r}(x) - \chi^t_{\breve{B}_s}(x)\right| + \left|\psi^t_{\breve{A}_r}(x) - \psi^t_{\breve{B}_s}(x)\right|\right)\right) \tag{6}$$

**Definition 16**. The Euclidean distance for two G-TSFSs $\breve{A}_r$ and $\breve{B}_s$ under each $x \in X$ is defined as:

$$E_1(\breve{A}_r, \breve{B}_s) = \frac{1}{2}\left(\left|r(x) - s(x)\right| + \sqrt{\frac{1}{2}\left((\phi^t_{\breve{A}_r}(x) - \phi^t_{\breve{B}_s}(x))^2 + (\chi^t_{\breve{A}_r}(x) - \chi^t_{\breve{B}_s}(x))^2 + (\psi^t_{\breve{A}_r}(x) - \psi^t_{\breve{B}_s}(x))^2\right)}\right) \tag{7}$$

**Definition 17**. The normalized Hamming distance for two G-TSFSs $\breve{A}_r$ and $\breve{B}_s$ in $X = \left\{x_1, x_2, ..., x_n\right\}$ is defined as:

$$H_2(\breve{A}_r, \breve{B}_s) = \frac{1}{2n}\left(\sum_{i=1}^{n}\left(\left|r(x_i) - s(x_i)\right| + \frac{1}{2}\left(\left|\phi^t_{\breve{A}_r}(x) - \phi^t_{\breve{B}_s}(x)\right| + \left|\chi^t_{\breve{A}_r}(x) - \chi^t_{\breve{B}_s}(x)\right| + \left|\psi^t_{\breve{A}_r}(x) - \psi^t_{\breve{B}_s}(x)\right|\right)\right)\right) \tag{8}$$

**Definition 18**. The normalized Euclidean distance for two G-TSFSs $\breve{A}_r$ and $\breve{B}_s$ in $X = \{x_1, x_2, ..., x_n\}$ is defined as:

$$E_2(\breve{A}_r, \breve{B}_s) = \frac{1}{2}\left( \frac{1}{n}\sum_{i=1}^{n} |r(x_i) - s(x_i)| + \sqrt{\frac{1}{2n}\sum_{i=1}^{n}\left( (\phi_{\breve{A}_r}^t(x_i) - \phi_{\breve{B}_s}^t(x_i))^2 + (\chi_{\breve{A}_r}^t(x_i) - \chi_{\breve{B}_s}^t(x_i))^2 + (\psi_{\breve{A}_r}^t(x_i) - \psi_{\breve{B}_s}^t(x_i))^2 \right)} \right) \quad (9)$$

**Example 3.** Consider the two G-TSFSs $\breve{A}$ and $\breve{B}$ in the universal set $X = \{x_1, x_2, x_3\}$ with the radiuses $r$ and $s$ assigned as {0.1, 0.23, 0.17} and {0.1, 0.07, 0.06}, respectively, that are detailed in Table 1. The illustration of these G-TSFSs in $X$ is presented in Fig. 3.

**Table 1.** The degrees of the element $x$

| $x \in X$ | $\breve{A}$ | | | | $\breve{B}$ | | | |
|---|---|---|---|---|---|---|---|---|
| | $\phi_{\breve{A}}$ | $\chi_{\breve{A}}$ | $\psi_{\breve{A}}$ | r | $\phi_{\breve{B}}$ | $\chi_{\breve{B}}$ | $\psi_{\breve{B}}$ | s |
| $x_1$ | 0.7 | 0.34 | 0.48 | 0.1 | 0.66 | 0.36 | 0.63 | 0.1 |
| $x_2$ | 0.7 | 0.37 | 0.51 | 0.23 | 0.68 | 0.38 | 0.55 | 0.07 |
| $x_3$ | 0.72 | 0.42 | 0.56 | 0.15 | 0.71 | 0.35 | 0.53 | 0.06 |

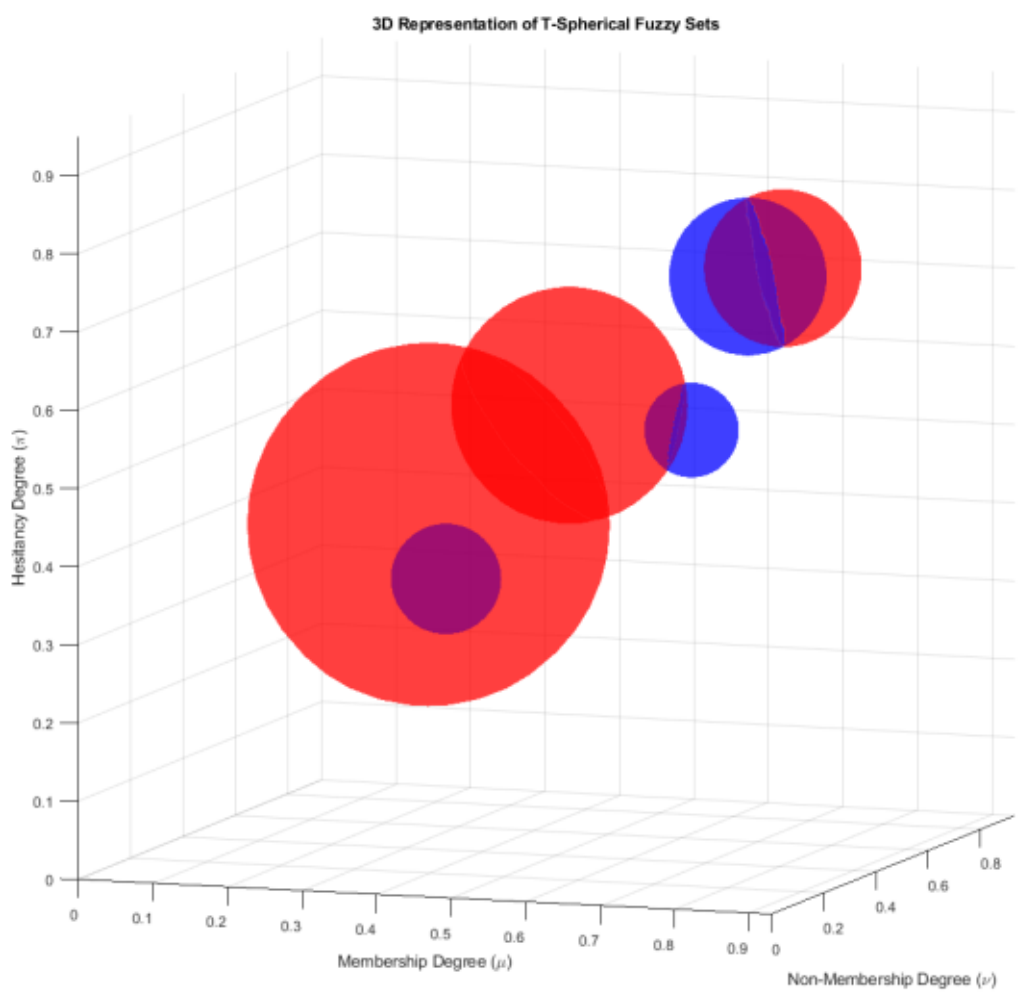

**Fig. 3**. Representation of $\breve{A}, \breve{B}$ in $X$ with $r = \{0.1, 0.23, 0.17\}$ and $s = \{0.1, 0.07, 0.06\}$

By implementing our proposed normalized Hamming distance and Euclidian distance defined in (8) and (9) to the above example, their respective results are shown in Table 2.

**Table 2.** Normalized Hamming and Euclidian distances

| Distance | Normalized Hamming $H_2$ | Normalized Euclidian $E_2$ |
|---|---|---|
| $D(\breve{A}_r, \breve{B}_s)$ | 0.07 | 0.239 |

## 4. Weighted Aggregation Operators

In this section, we introduce certain aggregation operators (AoPs) that incorporate weights into G-TSFVs. AoPs hold a significant position in supporting fuzzy decision-making systems. These operators are typically employed to gather and consolidate experts' information during various fuzzy analyses. The suggested operational principles aid in evaluating newly devised AoPs within the context of G-TSFVs. By employing addition and scalar multiplication operation principles, we can define the average AoPs. Similarly, combining multiplication and scalar power multiplication operational principles, we formulate geometric AoPs. The utilization of these developed AoPs assists in selecting the most optimal operational outcome for actual decision-making challenges in real-world situations. In the followings, we define some AoPs for G-TSFVs.

### 4.1. Globular T-Spherical Fuzzy (G-TSF) Weighted Average Aggregation Operator

**Definition 19**. Consider a family of G-TSFVs with $\left\{\alpha_i = \left\langle \phi_{\alpha_i}, \chi_{\alpha_i}, \psi_{\alpha_i}; r_{\alpha_i} \right\rangle : i = 1, 2, ..., n\right\}$. Then a G-TSF weighted average aggregation (G-TSFWAA) operator is defined as a mapping with

$$\text{G-TSFWAA}(\alpha_1, \alpha_2, ...\alpha_n) = \sum_{i=1}^{n} w_i \alpha_i \tag{10}$$

where $W = (w_1, w_2, ...w_n)$ is a weighted vector with $w_i \geq 0$ and $\sum_{i=1}^{n} w_i = 1$.

**Theorem 3**. Let a family of G-TSFVs be $a_{ri} = \left\langle \phi_{a_{ri}}, \chi_{a_{ri}}, \psi_{a_{ri}}; r_{a_{ri}} \right\rangle$ with $i = 1, 2, ..., n$. Then, G-TSFWAA can be written by using the propose operational laws, and converted as:

$$\text{G-TSFWAA}(a_{r1}, a_{r2}, ...a_{rn}) = \left\{ \sqrt[t]{1 - \prod_{i=1}^{n} (1 - \phi_{a_{ri}}^{t})^{w_i}}, \prod_{i=1}^{n} (\chi_{a_{ri}})^{w_i}, \prod_{i=1}^{n} (\psi_{a_{ri}})^{w_i}; \prod_{i=1}^{n} (r_{a_{ri}})^{w_i} \right\}.$$

Proof: We give the proof by employing the Mathematical Induction. Thus, it is as follows.

(a) Assume that i = 2. We know $w_1 a_{r1} = \left\{ \sqrt[t]{(1 - (1 - \phi_{a_{r1}}^{t})^{w_1}}, (\chi_{a_{r1}})^{w_1}, (\psi_{a_{r1}})^{w_1}; (r_{a_{r1}})^{w_1} \right\}$ and $w_2 a_{r2} = \left\{ \sqrt[t]{(1-(1-\phi_{a_{r2}}^t)^{w_2}}, (\chi_{a_{r2}})^{w_2}, (\psi_{a_{r2}})^{w_2}; (r_{a_{r2}})^{w_2} \right\}$. Thus, we have that G-TSFWAA$(a_{r1}, a_{r2})$

$$= w_1 a_{r1} + w_2 a_{r2} = \left\{ \sqrt[t]{(1-(1-\phi_{a_{r1}}^t)^{w_1}}, (\chi_{a_{r1}})^{w_1}, (\psi_{a_{r1}})^{w_1}; (r_{a_{r1}})^{w_1} \right\} + \left\{ \sqrt[t]{(1-(1-\phi_{a_{r2}}^t)^{w_2}}, (\chi_{a_{r2}})^{w_2}, (\psi_{a_{r2}})^{w_2}; (r_{a_{r2}})^{w_2} \right\}$$

$$= \left\{ \begin{array}{l} \sqrt[t]{((1-(1-\phi_{a_{r1}}^t)^{w_1}) + (1-(1-\phi_{a_{r2}}^t)^{w_2}) - (1-(1-\phi_{a_{r1}}^t)^{w_1})(1-(1-\phi_{a_{r2}}^t)^{w_2})}, \\ (\chi_{a_{r1}})^{w_1}.(\chi_{a_{r2}})^{w_2}, (\psi_{a_{r1}})^{w_1}.(\psi_{a_{r2}})^{w_2}; (r_{a_{r1}})^{w_1}.(r_{a_{r2}})^{w_2} \end{array} \right\}$$

$$= \left\{ \begin{array}{l} \sqrt[t]{1-(1-\phi_{a_{r1}}^t)^{w_1}(1-\phi_{a_{r2}}^t)^{w_2}}, \\ (\chi_{a_{r1}})^{w_1}.(\chi_{a_{r2}})^{w_2}, (\psi_{a_{r1}})^{w_1}.(\psi_{a_{r2}})^{w_2}; (r_{a_{r1}})^{w_1}.(r_{a_{r2}})^{w_2} \end{array} \right\}$$

$$= \left\{ \sqrt[t]{1-\prod_{i=1}^{2}(1-\phi_{a_{ri}}^t)^{w_i}}, \prod_{i=1}^{2}(\chi_{a_{ri}})^{w_i}, \prod_{i=1}^{2}(\psi_{a_{ri}})^{w_i}; \prod_{i=1}^{2}(r_{a_{ri}})^{w_i} \right\}$$

(b) Undertake that the outcome for $i=k$ is true, which is

$$\text{G-TSFWAA}(a_{r1}, a_{r2}, ... a_{rk}) = \left\{ \sqrt[t]{1-\prod_{i=1}^{k}(1-\phi_{a_{ri}}^t)^{w_i}}, \prod_{i=1}^{k}(\chi_{a_{ri}})^{w_i}, \prod_{i=1}^{k}(\psi_{a_{ri}})^{w_i}; \prod_{i=1}^{k}(r_{a_{ri}})^{w_i} \right\}$$

(c) Now, it is necessary to demonstrate the validity of the result for $i=k+1$ as follows.

$$\text{G-TSFWAA}(a_{r1}, a_{r2}, ... a_{rk}, a_{r(k+1)}) = \sum_{i=1}^{k} w_i a_{ri} + w_{k+1} a_{r(k+1)}$$

$$= \left\{ \sqrt[t]{1-\prod_{i=1}^{k}(1-\phi_{a_{ri}}^t)^{w_i}}, \prod_{i=1}^{k}(\chi_{a_{ri}})^{w_i}, \prod_{i=1}^{k}(\psi_{a_{ri}})^{w_i}; \prod_{i=1}^{k}(r_{a_{ri}})^{wi} \right\} + \left\{ \sqrt[t]{1-(1-\phi_{a_{r(k+1)}}^t)^{w_{k+1}}}, (\chi_{a_{r(k+1)}})^{w_{k+1}}, (\psi_{a_{r(k+1)}})^{w_{k+1}}; (r_{a_{r(k+1)}})^{w_{k+1}} \right\}$$

$$= \left\{ \begin{array}{l} \sqrt[t]{(1-\prod_{i=1}^{k}(1-\phi_{a_{ri}}^t)^{w_i}) + (1-(1-\phi_{a_{r(k+1)}}^t)^{w_{k+1}}) - (1-\prod_{i=1}^{k}(1-\phi_{a_{ri}}^t)^{w_i})(1-(1-\phi_{a_{r(k+1)}}^t)^{w_{k+1}})}, \\ \prod_{i=1}^{k}(\chi_{a_{ri}})^{w_i}.(\chi_{a_{r(k+1)}})^{w_{k+1}}, \prod_{i=1}^{k}(\psi_{a_{ri}})^{w_i}.(\psi_{a_{r(k+1)}})^{w_{k+1}}; \prod_{i=1}^{k}(r_{a_{ri}})^{wi}.(r_{a_{r(k+1)}})^{w_{k+1}} \end{array} \right\}$$

$$= \left\{ \begin{array}{l} \sqrt[t]{1-\prod_{i=1}^{k}(1-\phi_{a_{ri}}^t)^{w_i}(1-\phi_{a_{r(k+1)}}^t)^{w_{k+1}}}, \prod_{i=1}^{k}(\chi_{a_{ri}})^{wi}.(\chi_{a_{r(k+1)}})^{w_{k+1}}, \\ \prod_{i=1}^{k}(\psi_{a_{ri}})^{wi}.(\psi_{a_{r(k+1)}})^{w_{k+1}}; \prod_{i=1}^{k}(r_{a_{ri}})^{wi}.(r_{a_{r(k+1)}})^{w_{k+1}} \end{array} \right\}$$

$$= \left\{ \sqrt[t]{1-\prod_{i=1}^{k+1}(1-\phi_{a_{ri}}^t)^{w_i}}, \prod_{i=1}^{k+1}(\chi_{a_{ri}})^{w_i}, \prod_{i=1}^{k+1}(\psi_{a_{ri}})^{w_i}, \prod_{i=1}^{k+1}(r_{a_{ri}})^{wi} \right\}$$

Thus, for $i=k+1$, the result is satisfied. Consequently, the result is fulfilled for all $n$. ■

**Remark**: The G-TSFWAA operator evidently satisfies the following properties.

**(1) Idempotency**: Let a family of G-TSFVs be $a_{ri} = \langle \phi_{a_{ri}}, \chi_{a_{ri}}, \psi_{a_{ri}}; r_{a_{ri}} \rangle$ with $i = 1, 2, ..., n$. Then, the G-TSFWAA operator has idempotency, written as:

$$\text{G-TSFWAA}\,(a_{r1}, a_{r2}, \ldots. a_{rn}) = a_r\,.$$

**(2) Boundedness**: Let a family of G-TSFVs be $a_{ri} = \left\langle \phi_{a_{ri}}, \chi_{a_{ri}}, \psi_{a_{ri}}; r_{a_{ri}} \right\rangle$ with $i = 1, 2, \ldots, n$. Then, the G-TSFWAA operator has boundedness, written as:

$$a_{ri}^{-} \le \text{G-TSFWAA}(a_{r1}, a_{r2}, \ldots a_{rn}) \le a_{ri}^{+}$$

where $a_{ri}^{-} = \langle \min \phi_{a_{ri}}, \min \chi_{a_{ri}}, \max \psi_{a_{ri}}; \min r_{a_{ri}} \rangle$ and $a_{ri}^{+} = \langle \max \phi_{a_{ri}}, \min \chi_{a_{ri}}, \min \psi_{a_{ri}}; \min r_{a_{ri}} \rangle$.

**(3) Monotonicity**: Let a family of G-TSFVs be $a_{ri} = \left\langle \phi_{a_{ri}}, \chi_{a_{ri}}, \psi_{a_{ri}}; r_{a_{ri}} \right\rangle$ with $i = 1, 2, \ldots, n$. Then, the G-TSFWAA operator has monotonicity, written as:

If $a_{ri} \le \tilde{a}_{ri}$ for each $i = 1, 2, \ldots, n$, then $\text{G-TSFWAA}\,(a_{r1}, a_{r2}, \ldots a_{rn}) \le$ $\text{G-TSFWAA}\,(\tilde{a}_{r1}, \tilde{a}_{r2}, \ldots \tilde{a}_{rn})$.

### 4.2. G-TSF Weighted Geometric Aggregation Operator

**Definition 20**. Let $\left\{ \alpha_i = \left\langle \phi_{\alpha_i}, \chi_{\alpha_i}, \psi_{\alpha_i}; r_{\alpha_i} \right\rangle : i = 1, 2, \ldots, n \right\}$ be a family of G-TSFVs. The G-TSF weighted geometric aggregation (G-TSFWGA) operator is defined as a mapping with

$$\text{G-TSFWGA}\,(\alpha_1, \alpha_2, \ldots \alpha_n) = \prod_{i=1}^{n} w_i \alpha_i \tag{11}$$

where $W = (w_1, w_2, \ldots w_n)$ is a weighted vector with $w_i \ge 0$ and $\sum_{i=1}^{n} w_i = 1$.

**Theorem 4**. Let $\left\{ a_{ri} = \left\langle \phi_{a_{ri}}, \chi_{a_{ri}}, \psi_{a_{ri}}; r_{a_{ri}} \right\rangle : i = 1, 2, \ldots, n \right\}$ be a family of G-TSFVs. Then, G-TSFWGA can be written as:

$$\text{G-TSFWGA} = (a_{r1}, a_{r2}, \ldots a_{rn}) = \left\{ \prod_{i=1}^{n} (\phi_{a_{ri}})^{w_i}, \prod_{i=1}^{n} (\chi_{a_{ri}})^{w_i}, \sqrt[t]{1 - \prod_{i=1}^{n} (1 - \psi_{a_{ri}}^{t})^{w_i}}, \prod_{i=1}^{n} (r_{a_{ri}})^{w_i} \right\}.$$

Proof: The proof is similar as Theorem 3 based on the Mathematical Induction. ■

**Remark**: We can also get the properties of Idempotency, Boundedness and Monotonicity for the G-TSFWGA operator.

## 5. Formulation of Globular T-Spherical Fuzzy (G-TSF) MCGDM and Applications

In this section, we begin by introducing a similarity measure for our newly developed G-TSFSs. Subsequently, we leverage this measure to devise an innovative method to MCGDM scheme tailored to address decision-making challenges within the G-TSFS framework. We proceed to establish the practical utility and efficiency of our proposed G-TFS multi-criteria group decision making (G-TSF-MCGDM) scheme through its utilization in solving real-world decision-making problems.

### 5.1 A cosine similarity measure for G-TSFVs

Next, we suggest the implementation of a cosine similarity measure for G-TSFVs to determine their degree of similarity and resemblance.

**Definition 21**. Let $\tilde{e}_1 = \left\langle \overline{\phi}_{\tilde{e}_1}, \overline{\chi}_{\tilde{e}_1}, \overline{\psi}_{\tilde{e}_1}; r_{\tilde{e}_1} \right\rangle$ and $\tilde{e}_2 = \left\langle \overline{\phi}_{\tilde{e}_2}, \overline{\chi}_{\tilde{e}_2}, \overline{\psi}_{\tilde{e}_2}; r_{\tilde{e}_2} \right\rangle$ be two G-TSFVs. The cosine similarity measure (SM) $\tilde{S}_{CS}$ between the G-TSFVs $\tilde{e}_1$ and $\tilde{e}_2$ is defined as:

$$\tilde{S}_{CS}(\tilde{e}_1,\tilde{e}_2) = \tfrac{1}{2}\left\{ \frac{\overline{\phi}^t_{\tilde{e}_1}\overline{\phi}^t_{\tilde{e}_2} + \overline{\chi}^t_{\tilde{e}_1}\overline{\chi}^t_{\tilde{e}_2} + \overline{\psi}^t_{\tilde{e}_1}\overline{\psi}^t_{\tilde{e}_2}}{\sqrt{\overline{\phi}^{2t}_{\tilde{e}_1} + \overline{\chi}^{2t}_{\tilde{e}_1} + \overline{\psi}^{2t}_{\tilde{e}_1}}\sqrt{\overline{\phi}^{2t}_{\tilde{e}_2} + \overline{\chi}^{2t}_{\tilde{e}_2} + \overline{\psi}^{2t}_{\tilde{e}_2}}} + 1 - \left| r_{\tilde{e}_1} - r_{\tilde{e}_2} \right| \right\} \quad (12)$$

**Theorem 5.** Let two G-TSFVs be $\tilde{e}_1 = \left\langle \overline{\phi}_{\tilde{e}_1}, \overline{\chi}_{\tilde{e}_1}, \overline{\psi}_{\tilde{e}_1}; r_{\tilde{e}_1} \right\rangle$ and $\tilde{e}_2 = \left\langle \overline{\phi}_{\tilde{e}_2}, \overline{\chi}_{\tilde{e}_2}, \overline{\psi}_{\tilde{e}_2}; r_{\tilde{e}_2} \right\rangle$. Then, the proposed cosine SM $\tilde{S}_{CS}(\tilde{e}_1,\tilde{e}_2)$ satisfies the following features:

(i) $0 \le \tilde{S}_{CS}(\tilde{e}_1,\tilde{e}_2) \le 1$.

(ii) $\tilde{S}_{CS}(\tilde{e}_1,\tilde{e}_2) = \tilde{S}_{CS}(\tilde{e}_2,\tilde{e}_1)$.

(iii) $\tilde{S}_{CS}(\tilde{e}_1,\tilde{e}_2) = 1$, if $\tilde{e}_1 = \tilde{e}_2$.

Proof: Features (ii) and (iii) are straightly forward, and so we only prove Feature (i). As we have $r_{\tilde{e}_1}, r_{\tilde{e}_2} \in [0,1]$, it is obvious that the term $1 - \left| r_{\tilde{e}_1} - r_{\tilde{e}_2} \right|$ will be within the unit interval. *i.e.* $0 \le 1 - \left| r_{\tilde{e}_1} - r_{\tilde{e}_2} \right| \le 1$. Whereas, the other mathematical expression within the cosine similarity measure $\tilde{S}_{CS}(\tilde{e}_1,\tilde{e}_2)$ is $\dfrac{\overline{\phi}^t_{\tilde{e}_1}\overline{\phi}^t_{\tilde{e}_2} + \overline{\chi}^t_{\tilde{e}_1}\overline{\chi}^t_{\tilde{e}_2} + \overline{\psi}^t_{\tilde{e}_1}\overline{\psi}^t_{\tilde{e}_2}}{\sqrt{\overline{\phi}^{2t}_{\tilde{e}_1} + \overline{\chi}^{2t}_{\tilde{e}_1} + \overline{\psi}^{2t}_{\tilde{e}_1}}\sqrt{\overline{\phi}^{2t}_{\tilde{e}_2} + \overline{\chi}^{2t}_{\tilde{e}_2} + \overline{\psi}^{2t}_{\tilde{e}_2}}}$ that just represents the cosine value corresponding to a particular angle within a range of $[0, \frac{\pi}{2}]$ characterized by the G-TSFVs $\tilde{e}_1$ and $\tilde{e}_2$. Thus, it would be lying within the unit interval [0,1]. Hence, it completes the proof, *i.e.* $0 \le \tilde{S}_{CS}(\tilde{e}_1,\tilde{e}_2) \le 1$. ■

### 5.2 A G-TSF-MCGDM Scheme

Multi-criteria group decision making (MCGDM) in a fuzzy surrounding is a process aimed at making decisions in situations where there are multiple criteria to be considered in which uncertainty exists due to imprecise or vague information. In such scenarios, traditional decision-making approaches may struggle to adequately capture the problem's intricate and ambiguous nature. Fuzzy logic provides a powerful structure for managing ambiguity by allowing for the representation of vague concepts and the incorporation of qualitative judgments. In MCGDM, the input of multiple decision makers with diverse perspectives and expertise is taken into account, leading to more comprehensive and robust decision outcomes. In this subsection, we construct a MCGDM scheme using our proposed cosine SM tailored for our newly introduced G-TSFVs. To do so, we initially establish a theoretical framework delineating the implementation procedure of our proposed scheme for addressing MCGDM problems under G-TSF environment. Let us consider a set of *m* options $\xi = \{\xi_1, \xi_2, ..., \xi_m\}$ to be evaluated by a group of *l* informed decision makers $\varepsilon = \langle \varepsilon_1, \varepsilon_2, ..., \varepsilon_l \rangle$ based on predefined *n* attributes $\zeta = \{\zeta_1, \zeta_2, ..., \zeta_n\}$ . The assessment of the $p^{th}$ option with respect to the $q^{th}$ attribute is represented by using the G-TSFVs $G_{pq} = \langle \zeta_q, \phi_{\xi_p}(\zeta_q), \chi_{\xi_p}(\zeta_q), \psi_{\xi_p}(\zeta_q); r_{\xi_p}(\zeta_q) \rangle$, $p = 1, 2, ..., m$ , $q = 1, 2, ..., n$ Consequently, the overall assessments provided by decision makers for the *m* options $\xi = \{\xi_1, \xi_2, ..., \xi_m\}$ across the *n* attributes $\zeta = \{\zeta_1, \zeta_2, ..., \zeta_n\}$ are formulated into a G-TSF decision making with $G = \langle G_{pq} \rangle_{m \times n}$ .

This matrix serves to express the value of each option with regards to each attribute from the perspective of decision makers. Here, $G_{pq}$ shows the importance of the $p^{th}$ alternative $\xi_p$ on the basis of the $q^{th}$ attribute $\zeta_q$ in the perspective of the decision makers. In order to identify the ideal substitute, it is imperative to rank the options by comparing them against an ideal benchmark, considering their strength and weaknesses across attributes. To construct this ideal benchmark, we employ the concept of Positive Ideal Solution (PIS). An alternative in terms of G-TSFVs is said to be ideal choice if it has maximum value of DoM and maximum possible value of radius while having minimum DoI and the value of DoN. Thus, we extend the notion of PIS to serve as an ideal alternative for G-TSFVs, establishing the theoretical foundation of our Ideal Alternative (IA) for G-TSFVs as:

$$\xi^{*}=\left\{\left\langle \zeta_{q},\phi_{\xi}^{*}\left(\zeta_{q}\right),\chi_{\xi}^{*}\left(\zeta_{q}\right),\psi_{\xi}^{*}\left(\zeta_{q}\right);r_{\xi}^{*}\left(\zeta_{q}\right)\right\rangle :\zeta_{q}\in\zeta\right\},q=1,2,...,n. \tag{13}$$

Here, $\phi_{\xi}^{*}\left(\zeta_{q}\right)=1$, $\chi_{\xi}^{*}\left(\zeta_{q}\right)=0$, $\psi_{\xi}^{*}\left(\zeta_{q}\right)=0$, and $r_{\xi}^{*}\left(\zeta_{q}\right)=1$ for $p=1,2,...,m$ and $q=1,2,...,n.$

Next, we apply our scheme to compute the similarity between the available choices and the IA using our proposed cosine SM $\tilde{S}_{CS}$ between two G-TSFVs outlined in Eq. (12) with

$$\tilde{S}_{CS}(\xi_{p},\xi^{*})=\tfrac{1}{2}\left\{\frac{1}{n}\sum_{q=1}^{n}\frac{\bar{\phi}_{\xi_{p}}^{t}\left(\zeta_{q}\right)}{\sqrt{\bar{\phi}_{\xi_{p}}^{2t}\left(\zeta_{q}\right)+\bar{\chi}_{\xi_{p}}^{2t}\left(\zeta_{q}\right)+\bar{\psi}_{\xi_{p}}^{2t}\left(\zeta_{q}\right)}}+r_{\xi_{p}}\left(\zeta_{q}\right)\right\} \tag{14}$$

A higher similarity value indicates a better choice. Subsequently, we select the option with the highest degree of similarity to the IA as the top choice, adhering to the principle of maximum similarity. This is formalized in the mathematical expression of our proposed G-TSF-MAGDM scheme as:

$$\tilde{S}_{CS}^{*}=\max_{1\leq p\leq m}\tilde{S}_{CS}(\xi_{p},\xi^{*}) \tag{15}$$

Therefore, the $p^{th}$ alternative, exhibiting the maximum similarity to the IA compared to other available options, is designated as the best alternative. Moreover, we present the useful application of our proposed scheme by solving a genuine MCGDM problem in the followings. This showcase illustrates the relevance and practicality of our approach in practical decision-making circumstances.

**Example 4.** In the contemporary fast-paced and competitive landscape, the significance of skill development has risen as a crucial element for both organizations and educational institutions in order to sustain a competitive advantage. Acknowledging the value of ongoing learning and advancement, a skill-development company has made the effort to offer extensive instruction plans customized for staffs spanning various educational institutions. Central to the success of these programs is the selection of an optimal venue. The skill development organization seeks to establish an environment conducive to learning, one that not only enhances participants' knowledge and abilities but also ignites their drive for excellence. The venue selection process prioritizes several key characteristics, each essential in delivering an exceptional training experience. To guarantee the efficacy of an upcoming training session, the organization recognizes the importance of conducting a comprehensive market analysis. This questionnaire seeks to pinpoint and narrow down the top four probable venues according to the top five crucial qualities. To streamline this process, the association has formed a devoted committee consisting of field specialists. Utilizing their expertise, these professionals will thoroughly assess the possibilities that are accessible. The committee has identified the following five individualities as paramount in selecting the ideal location for the training sessions.

One of the primary deliberations lies in the venue's site and its natural surroundings. Electing for a setting that bids picturesque views, serene environments, and a peaceful atmosphere is paramount. By selecting such a location, the organization strives to cultivate an environment fostering creativity, concentration, and stimulus. Additionally, ability stands as another crucial aspect when choosing the location. Confirming that the location of choice can adequately accommodate the anticipated number of contributors is essential for preserving an optimal education atmosphere. Adequate facilitating engaging sessions and hands-on activities requires strategically arranged training rooms, comfortable seating, and appropriate equipment. Moreover, the facilities and amenities offered by the site play a pivotal role in providing a flawless instruction environment. All of these elements contemporary video technology, trustworthy Wi-Fi connectivity, cozy breakout spaces, discussion areas with tiny libraries, and luxurious refreshment areas help make instruction sessions more convenient and productive overall. The atmosphere and setting of the location are crucial factors in creating a conducive learning atmosphere significantly influence the learning experience. Elements like appropriate lighting, sound management, temperature regulation, and hygiene maintenance are meticulously assessed. To guarantee that participants are able to engage with the preparation gratified without any interruptions, thereby fostering an atmosphere conducive to learning and knowledge preoccupation. Furthermore, practical support emerges as another crucial facet to ensure the flat execution of the instruction involves. The selected scene should have committed technical staff available to deal promptly with any technical problems that may arise and to ensure that the smooth provision of presentations using many media and other interactive elements throughout the instructional process, guaranteeing seamless experience sessions. A symbolic depiction elucidating these essential physiognomies is presented as follows: (i) Location and natural beauty surrounding the venue; (ii) Capacity; (iii) Amenities and facilities; (iv) Amenities and environment; (v) Technical support.

The nominated top four venues and the essential features they embody are represented in set notation as $\nu = \langle \nu_1, \nu_2, \nu_3, \nu_4 \rangle$ and $f = \langle f_1, f_2, f_3, f_4, f_5 \rangle$, respectively. Through a rigorous assessment process, the committee of three field specialists $\varepsilon = \langle \varepsilon_1, \varepsilon_2, \varepsilon_3 \rangle$ have meticulously analyzed and deliberated upon each alternative $\nu = \langle \nu_1, \nu_2, \nu_3, \nu_4 \rangle$ based on predefined features $f = \langle f_1, f_2, f_3, f_4, f_5 \rangle$, utilizing the notion of G-TSFVs to represent their decisions. The culmination of their evaluations is initially presented in a comprehensive and concise T-spherical fuzzy decision matrix (TSF-DM) in the form of T-spherical fuzzy values (TSFVs). This matrix encapsulates the expert insights and considerations for each of the potential venues, as illustrated in Table 3 serving as a valuable resource; the TSF-DM equips the skill development organization with the necessary information to make an informed and confident decision regarding the selection of the most suitable venue for their training sessions.

Table 3. T-spherical fuzzy decision matrix

| Evaluators | Options | Features ( $f$ ) | | | | |
|---|---|---|---|---|---|---|
| | | $f_1$ | $f_2$ | $f_3$ | $f_4$ | $f_5$ |
| $\varepsilon_1$ | $\nu_1$ | <0.73,0.44,0.54> | <0.69,0.35,0.43> | <0.83,0.32,0.56> | <0.76,0.62,0.53> | <0.85,0.38,0.56> |
| | $\nu_2$ | <0.80,0.28,0.56> | <0.68,0.42,0.62> | <0.63,0.40,0.53> | <0.70,0.40,0.60> | <0.88,0.46,0.53> |
| | $\nu_3$ | <0.9,0.22,0.48> | <0.83,0.3,0.35> | <0.88,0.35,0.42> | <0.9,0.2,0.4> | <0.8,0.34,0.42> |
| | $\nu_4$ | <0.76,0.43,0.50> | <0.69,0.38,0.49> | <0.8,0.4,0.5> | <0.78,0.3,0.5> | <0.76,0.38,0.48> |
| $\varepsilon_2$ | $\nu_1$ | <0.64,0.36,0.55> | <0.76,0.42,0.53> | <0.74,0.42,0.52> | <0.72,0.28,0.63> | <0.8,0.4,0.5> |
| | $\nu_2$ | <0.76,0.43,0.52> | <0.63,0.3,0.7> | <0.7,0.4,0.6> | <0.73,0.37,0.53> | <0.83,0.4,0.5> |
| | $\nu_3$ | <0.88,0.26,0.46> | <0.9,0.28,0.39> | <0.80,0.29,0.50> | <0.84,0.36,0.52> | <0.92,0.18,0.46> |
| | $\nu_4$ | <0.8,0.5,0.6> | <0.76,0.4,0.6> | <0.69,0.37,0.53> | <0.70,0.40,0.60> | <0.74,0.43,0.52> |
| $\varepsilon_3$ | $\nu_1$ | <0.58,0.28,0.46> | <0.65,0.33,0.49> | <0.53,0.38,0.46> | <0.68,0.36,0.52> | <0.72,0.43,0.6> |
| | $\nu_2$ | <0.65,0.39,0.6> | <0.68,38,0.57> | <0.72,0.36,0.53> | <0.71,0.28,0.48> | <0.74,0.31,0.54> |
| | $\nu_3$ | <0.93,0.15,0.40> | <0.88,0.35,0.42> | <0.86,0.33,0.46> | <0.86,0.40,0.50> | <0.87,0.26,0.40> |
| | $\nu_4$ | <0.9,0.3,0.45> | <0.8,0.3,0.7> | <0.72,0.4,0.51> | <0.75,0.29,0.54> | <0.48,0.5,0.4> |

Utilizing the results of our proposed G-TSFVs in Theorem 1, that was also outlined in Eq. (2), we transform the values of TSFVs presented in Table 3 for each alternative, along with the corresponding average values of DOM, DOI and DoN. These G-TSFVs' membership grads are elucidated in Table 4.

Table 4. Average TSF-DM of the evaluation values by experts

| Options | Feature $f$ | | | | |
|---|---|---|---|---|---|
| | $f_1$ | $f_2$ | $f_3$ | $f_4$ | $f_5$ |
| $v_1$ | <0.65,0.36,0.51> | <0.7,0.36,0.48> | <0.7,0.37,0.51> | <0.72,0.42,0.56> | <0.79,0.40,0.55> |
| $v_2$ | <0.73,0.39,0.56> | <0.66,0.36,0.63> | <0.68,0.38,0.55> | <0.71,0.35,0.53> | <0.81,0.39,0.52> |
| $v_3$ | <0.90,0.21,0.44> | <0.87,0.31,0.38> | <0.84,0.32,0.46> | <0.86,0.32,0.47> | <0.86,0.26,0.42> |
| $v_4$ | <0.82,0.41,0.51> | <0.75,0.36,0.59> | <0.73,0.39,0.51> | <0.74,0.33,0.54> | <0.66,0.43,0.46> |

We calculate the radius of the sphere presented in Theorem 1 by utilizing the average values of (DoM), (DoI) and (DoN) for each alternative shown in Table 4, along with our proposed method for computing the radius of the sphere $r$ as provided in Eq. (3). The maximum radii for each alternative, expressed in terms of G-TSFVs, are then presented in Table 5.

Table 5. Summary of maximum radii of the spheres of the alternatives

| Options | Feature $f$ | | | | |
|---|---|---|---|---|---|
| | $f_1$ | $f_2$ | $f_3$ | $f_4$ | $f_5$ |
| $v_1$ | 0.11 | 0.10 | 0.23 | 0.15 | 0.13 |
| $v_2$ | 0.14 | 0.10 | 0.08 | 0.06 | 0.15 |
| $v_3$ | 0.07 | 0.09 | 0.10 | 0.09 | 0.14 |
| $v_4$ | 0.13 | 0.08 | 0.07 | 0.10 | 0.20 |

Next, we integrate the evaluation values of the alternatives outlined in both Table 3 and Table 4 to construct the globular T-spherical decision matrix (G-TSF-DM). The resultant G-TSF-DM is provided in Table 6. In order to identify the best alternative, we establish a theoretical benchmark using our proposed ideal alternative (IA) described in Eq. (13) as

$$v^* = \left\{ \left\langle f_q, 1, 0, 0; 1 \right\rangle : f_q \in f \right\}, q = 1, 2, ..., 5. \tag{16}$$

Table 6. G-TSF decision matrix

| Options | Feature $f$ | | | | |
|---|---|---|---|---|---|
| | $f_1$ | $f_2$ | $f_3$ | $f_4$ | $f_5$ |
| $\nu_1$ | <0.65,0.36,0.51;0.11> | <0.7,0.36,0.48;0.10> | <0.7,0.37,0.51;0.23> | <0.72,0.42,0.56;0.15> | <0.79,0.40,0.55;0.13> |
| $\nu_2$ | <0.73,0.39,0.56;0.14> | <0.66,0.36,0.63;0.10> | <0.68,0.38,0.55;0.08> | <0.71,0.35,0.53;0.06> | <0.81,0.39,0.52;0.15> |
| $\nu_3$ | <0.90,0.21,0.44;0.07> | <0.87,0.31,0.38;0.09> | <0.84,0.32,0.46;0.10> | <0.86,0.32,0.47;0.09> | <0.86,0.26,0.42;0.14> |
| $\nu_4$ | <0.82,0.41,0.51;0.13> | <0.75,0.36,0.59;0.08> | <0.73,0.39,0.51;0.07> | <0.74,0.33,0.54;0.10> | <0.66,0.43,0.46;0.20> |

Upon applying our devised cosine similarity measure $\tilde{S}_{CS}$, as defined in Eq. (12), we assess the likeness of the optimal alternative and each of the available options. The outcomes of this analysis are detailed in Table 7. Out of the four venues currently under consideration, the optimal alternative for conducting future training sessions can be determined by selecting the option with the highest similarity value, utilizing our proposed similarity measure based on G-TSF-MCGDM.

Table 7. Similarity of IA with alternatives under study

| Options ($\nu_p$) | $\tilde{S}_{CS}(\nu^*, \nu_p)$ |
|---|---|
| $\nu_1$ | 0.5299 |
| $\nu_2$ | 0.4931 |
| $\nu_3$ | **0.5447** |
| $\nu_4$ | 0.5257 |

Table 7 reveals that the 3rd alternative, specifically $\nu_3$ (i.e. Venue 3) stands out as the option with the highest degree of similarity to the IA among all four venues, according to our proposed similarity measure. Consequently, selecting the 3rd Venue is the optimal choice for effectively conducting the training session, ensuring a highly conducive and outcome-oriented teaching and learning environment. Additionally, we have ranked all four venues based on their similarities to the ideal choice $\nu^*$ as $\nu_3 > \nu_1 > \nu_4 > \nu_2$.

## 6 Conclusion

In this paper, we present the innovative idea of G-TSFSs, expanding upon existing frameworks such as IFSs, PyFSs, C-IFSs, C-PyFSs, SFSs, C-SFSs, and TSFSs. The G-TSFS model provides a refined representation of DoM, DoI and DoN using a sphere, thereby offering a more precise depiction of vague, ambiguous, and imprecise information. By structuring data points on a sphere with defined center and radius parameters, this G-TSFS model significantly enhances decision-making processes by enabling a comprehensive evaluation of objects within a flexible region. Building upon the definition of G-TSFSs with theorems, we also establish fundamental set operations and introduces algebraic operations tailored for G-TSFVs. These activities augment the evaluative capabilities of decision-makers, facilitating more nuanced decision-making processes across a broader spectrum of scenarios. Further, we propose a radius-based similarity measure to quantify the similarity between G-TSFVs as well as Hamming and Euclidean distance measures specific to G-TSFVs. Theoretical theorems and practical numerical examples are provided to elucidate the computational mechanisms underlying these measures. Additionally, we give the two aggregation operators, G-TSFWAA and G-TSFWGA, designed specifically for G-TSFVs. We also developed a MCGDM scheme for G-TSFSs, termed G-TSF-MCGDM, to address MCGDM challenges effectively. The applicability and efficiency of the proposed G-TSF-MCGDM method are established through its application to solve the selection problem of the best venue for professional development training sessions in a firm. The analysis results authenticate the suitability and utility of the proposed technique for resolving MCGDM problems, affirming its effectiveness in practical decision-making scenarios across various domains.